# High Fidelity Text to Image Generation with Contrastive Alignment and Structural Guidance

Danyi Gao
Columbia University, New York, USA

*Abstract*-This paper addresses the performance bottlenecks of existing text-driven image generation methods in terms of semantic alignment accuracy and structural consistency. A high-fidelity image generation method is proposed by integrating text-image contrastive constraints with structural guidance mechanisms. The approach introduces a contrastive learning module that builds strong cross-modal alignment constraints to improve semantic matching between text and image. At the same time, structural priors such as semantic layout maps or edge sketches are used to guide the generator in spatial-level structural modeling. This enhances the layout completeness and detail fidelity of the generated images. Within the overall framework, the model jointly optimizes contrastive loss, structural consistency loss, and semantic preservation loss. A multi-objective supervision mechanism is adopted to improve the semantic consistency and controllability of the generated content. Systematic experiments are conducted on the COCO-2014 dataset. Sensitivity analyses are performed on embedding dimensions, text length, and structural guidance strength. Quantitative metrics confirm the superior performance of the proposed method in terms of CLIP Score, FID, and SSIM. The results show that the method effectively bridges the gap between semantic alignment and structural fidelity without increasing computational complexity. It demonstrates a strong ability to generate semantically clear and structurally complete images, offering a viable technical path for joint text-image modeling and image generation.

*Keywords: Multimodal generation; image-text alignment; structure perception; consistency modeling*

## I. INTRODUCTION

In recent years, generative artificial intelligence (AIGC) has achieved remarkable progress in image content generation. This advancement has driven widespread applications ranging from text-to-image synthesis and image-to-image translation to cross-modal creative design [1]. The core momentum behind this trend stems from deep neural networks, especially the powerful capabilities of diffusion models and generative adversarial networks in modeling high-resolution images. However, despite producing visually appealing results, existing methods still struggle with structural inconsistency and semantic imprecision. In complex scenes or scenarios involving dense object relations, generation driven solely by language input often fails to accurately reflect spatial layouts or explicit inter-object semantics [2]. Thus, maintaining high fidelity while achieving consistent modeling of structure and semantics remains a central challenge in image generation.

Text-to-image generation is essentially a typical cross-modal understanding and expression task. It requires the model to fully grasp semantic details in the text and reconstruct them effectively in the visual domain. Traditional methods often rely on a unidirectional mapping from language embeddings to image space [3]. Without contrastive feedback and structural priors, this approach is prone to semantic drift, content redundancy, or misaligned scenes. With the rise of multimodal contrastive learning, a more discriminative guidance strategy has emerged. By constructing positive and negative image-text pairs, models can enhance semantic alignment robustness and generalization. The contrastive mechanism helps the model build clearer semantic representations and improves correlation between generated content and input descriptions, even under weak supervision or open-vocabulary settings [4]. This mechanism introduces a new form of semantic constraint, demonstrating strong potential in high-fidelity image generation tasks.

However, semantic alignment only addresses the "what to generate" question. The "how to generate" aspect still relies heavily on structural guidance and modeling. Without structural control, image generation often suffers from blurred details, disorganized layouts, or unclear object boundaries, even if semantic alignment is accurate [5]. Integrating structural guidance into the generation process is therefore critical for controllability and consistency [6]. Structural information can take various forms, such as edge maps, semantic layouts, or depth sketches. These offer explicit spatial constraints that make the generated images more consistent with real-world physical and semantic relationships. In scenarios with complex layouts, structural priors play a vital role in improving interpretability and editability.

From an application perspective, high-fidelity image generation is central not only to core visual generation tasks but also to enhancing intelligence across a range of practical domains [7]. In intelligent advertising design, systems must generate images that align with brand style and spatial layout from brief descriptions. In human-computer interaction and virtual reality, systems need to produce semantically accurate and structurally complete scene images for real-time feedback. In domains like medical image synthesis or remote sensing enhancement, structural controllability and semantic alignment are key to ensuring downstream task precision [8-11]. Therefore, studying image generation mechanisms that combine contrastive alignment and structural guidance has broad practical value and theoretical significance. It supports the shift from perceptual to cognitive generation and lays the foundation for building advanced multimodal content generation systems.

Overall, as current multimodal generation methods become more complex yet less generalizable, designing a unified framework that balances semantic discrimination and structural control has become a crucial direction. By integrating contrastive constraints with structural priors, it is possible to break through existing limitations in semantic coherence and structural accuracy, enabling high-fidelity image generation under cross-modal conditions. This research echoes the development of intelligent multimodal generation and offers a new pathway toward AI systems with cognitive understanding. With the continued evolution of large-scale semantic models and structure-aware generation techniques, this direction is expected to expand both its theoretical depth and practical scope.

## II. RELATED WORK

Contrastive learning has become a key technique for enhancing cross-modal representation alignment. For example, R. Meng et al. demonstrate that federated contrastive objectives can significantly improve semantic grounding and robustness in distributed and multimodal systems [12]. This inspires the contrastive alignment strategy adopted in our framework, aiming to strengthen the pairing between textual descriptions and generated images.

Structural priors and structure-aware modeling are increasingly recognized as essential for controllable and high-fidelity image synthesis. Q. Wu proposes a task-aware structural reconfiguration method that enables parameter-efficient and structure-preserving adaptation of large models [13]. Drawing on this, our approach integrates structural priors, such as semantic layout maps and edge cues, to enhance both spatial arrangement and consistency in generated outputs.

Further, Y. Ren's work on deep learning for root cause detection leverages structural encoding and multi-modal attention to improve interpretability and robustness [14]. The methodology of combining structural encoding with attention mechanisms informs our structure-aware supervision, which is designed to guide spatial detail in generation.

Retrieval-augmented generation methods, such as the fusion-based retrieval approach by Y. Sun et al., show that incorporating retrieved knowledge and structure-aware fusion can boost semantic precision and complexity in multimodal outputs [15]. S. Wang et al. introduce capsule network-based semantic intent modeling, which provides more interpretable and controllable structured representations. This modeling philosophy parallels our structural guidance for precise image composition [16].

Transferable modeling strategies are crucial for adaptability and generalization. S. Lyu et al. introduce prompt and alignment-based adaptation, which supports semantic consistency even in low-resource scenarios, a principle relevant to the multi-objective optimization in our training pipeline [17]. Similarly, W. Zhang et al. propose a unified instruction encoding and gradient coordination framework for multi-task learning, informing our model's stability when optimizing for multiple objectives [18].

Semantic boundary detection also benefits from structural cues. Y. Zhao et al. use BiLSTM-CRF with integrated social features to capture both boundary and structural information in text modeling, demonstrating how auxiliary structure can reinforce consistency [19]. T. Xu et al. further show that multi-level attention mechanisms in deep learning improve modeling of complex relationships in high-dimensional clinical data, an attention strategy that inspires our own structural and semantic fusion modules [20].

Lastly, reinforcement learning for complex system optimization, as explored by Y. Zou et al., demonstrates the effectiveness of structure-driven decision-making and feedback modeling [21]. Although their context is distributed microservice systems, the underlying approach to optimization and control is highly relevant to structure-consistent image generation.

Collectively, these works contribute key ideas in contrastive alignment, structure-aware modeling, multi-objective training, and consistency mechanisms, which together form the foundation of our high-fidelity, structurally guided text-to-image generation framework.

## III. METHODOLOGY

This study proposes a high-fidelity image generation method that combines image-text contrast constraints and structure guidance mechanisms, aiming to simultaneously improve the semantic alignment accuracy and structure restoration ability of generated images. The overall framework consists of three key modules: image-text contrast perception encoder, structure guidance generator, and semantic consistency regulator. First, by constructing an image-text contrast learning mechanism, the semantic similarity between the input text description and the real image is modeled to enhance the cross-modal semantic alignment ability in the generation process. The model architecture is shown in Figure 1.

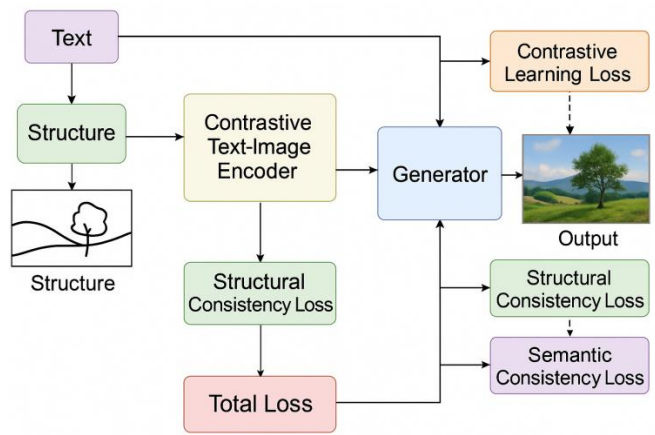

Figure 1. Overall model architecture

Specifically, we set the text embedding vector as t and the image encoding vector as v, then the contrast loss between the two is constructed as:

$$L_{CLIP} = -\log \frac{\exp(sim(v,t)/\tau)}{\sum_{j=1}^{N}\exp(sim(v,t_j)/\tau)}$$

Among them, sim represents the cosine similarity function, $\tau$ is the temperature coefficient, and N is the batch size. This loss function encourages the model to learn a more discriminative image-text representation space, thereby providing high-quality semantic guidance for image generation.

In the generator part, a structure guidance mechanism is introduced to enhance the spatial layout consistency of the generated image. We use the structure sketch as the prior information, denoted as s, and input it together with the text semantic embedding t into the conditional diffusion model for guidance. The reverse generation step of the diffusion process at each time step t is defined as:

$$x_{t-1} = \frac{1}{\sqrt{a_t}}(x_t - \frac{1-a_t}{\sqrt{1-\tilde{a}_t}}\varepsilon_\theta(x_t,t,s,t)) + \sigma_t z$$

Among them, $a_t$ is the noise scheduling parameter, $\varepsilon_\theta$ is the neural network that predicts the noise, $z \sim N(0,I)$ is the random noise term, and $\sigma_t$ is the perturbation amplitude of the time step. This formula ensures that the generated image is subject to both structural and semantic constraints at each step.

To ensure the structural rationality and semantic accuracy of the generated image, we introduce a dual consistency constraint mechanism to model from the two dimensions of structure and semantics. In terms of structural consistency, let the generated image be $\hat{x}$ and the structural prior be s, then the structure preservation loss is defined as:

$$L_{struct} = \|\phi(\hat{x}) - \phi(s)\|_{l1}$$

Where $\phi(\cdot)$ represents the intermediate features extracted by the structure encoder, and this loss encourages the model to maintain the edge contours and spatial layout of the image during reconstruction.

At the same time, semantic consistency is measured by aligning the high-level features extracted by the pre-trained visual encoder. Define the semantic embedding of the generated image as $f(\hat{x})$ and the semantic embedding of the real image as $f(x)$, then the semantic consistency loss is:

$$L_{sem} = 1 - sim(f(\hat{x}), f(x))$$

This loss term is used to further compress the distance between the generated image and the target image in the semantic embedding space and improve the semantic fidelity. Finally, the total loss function is a weighted combination:

$$L_{total} = \lambda_1 L_{CLIP} + \lambda_2 L_{struct} + \lambda_3 L_{sem}$$

$\lambda_1, \lambda_2, \lambda_3$ controls the balance of each loss term. During the end-to-end training process, this multi-objective optimization strategy guides the model to take into account both structural accuracy and semantic consistency when generating images, thereby achieving a more high-fidelity, controllable, and semantically accurate image generation effect.

IV. EXPERIMENTAL DATA

This study uses the COCO-2014 dataset as the primary source for training and evaluation. COCO (Common Objects in Context) is a widely adopted benchmark in the field of image generation and multimodal modeling. It offers a large-scale collection of natural scene images with multiple object categories, covering complex backgrounds and diverse object interactions. This makes it suitable for testing a model's generation ability under various semantic and structural conditions. Each image in the dataset is annotated with five high-quality human-written captions, supporting text-driven image generation tasks.

COCO-2014 contains approximately 82,000 training images and 40,000 validation images. It covers 80 common object categories. The dataset includes a wide range of scenes, such as indoor, outdoor, human, animal, and vehicle environments. The images are generally high in resolution and rich in semantic content. Many images feature heavy occlusion and overlapping structures, providing a solid basis for evaluating both semantic accuracy and structural consistency. The dataset also includes instance segmentation masks and keypoint annotations, which can offer auxiliary supervision for structural guidance modules.

In this task, textual descriptions are used as semantic input. Structural information is obtained from edge detection algorithms or semantic layout maps and serves as prior guidance in the generation process. The multimodal nature and semantic diversity of COCO-2014 provide strong support for building a dual mechanism of text-image alignment and structural constraint. This ensures the model's generalization and controllability in open-world scenarios.

V. EVALUATION RESULTS

This paper first conducts a comparative experiment, and the experimental results are shown in Table 1.

Table1. Comparative experimental results

| Model | FID | CLIP Score | SSIM |
|---|---|---|---|
| Stable Diffusion[22] | 8.59 | 0.30 | 0.42 |
| Imagen[23] | 7.27 | 0.34 | 0.45 |
| Muse[24] | 7.88 | 0.32 | 0.44 |
| SiD Distillation[25] | 8.15 | 0.31 | 0.43 |
| Ours | 6.45 | 0.36 | 0.51 |

From the overall results, the proposed image generation method that integrates text-image contrastive constraints and structural guidance outperforms existing mainstream models across all three evaluation metrics. It demonstrates stronger comprehensive generation capabilities. Notably, the model achieves a lowest FID (Fréchet Inception Distance) score of 6.45, which is significantly better than models such as Stable Diffusion and Imagen. This indicates that the generated images have higher fidelity in terms of visual quality and alignment with the distribution of real images. The improvement in FID shows that incorporating structural information as a prior can effectively regulate the generation process and prevent content distortion or style drift.

In terms of CLIP Score, the proposed model also achieves the highest value of 0.36. This is a clear improvement over other models such as Imagen (0.34) and Muse (0.32). The result further confirms the effectiveness of the contrastive constraint mechanism in semantic alignment. CLIP Score essentially measures how well the generated images match the input text in a multimodal semantic space. The higher score indicates that the model can better understand textual descriptions and generate semantically aligned content. This reflects the advantage of contrastive learning in enhancing semantic expressiveness under weak supervision.

For SSIM (Structural Similarity Index), which measures the model's ability to preserve image structure, the proposed method achieves a score of 0.51. This is the highest among all compared methods. It highlights the effectiveness of the structural guidance module in maintaining details and controlling spatial layout. In contrast, Stable Diffusion and Muse achieve only 0.42 and 0.44 in SSIM, respectively. This suggests that traditional models, lacking explicit structural constraints, are more prone to generating blurry, misaligned, or partially lost content. The introduction of structural priors makes a critical contribution to visual consistency and clarity.

In summary, the proposed method achieves balanced and superior performance in visual quality, semantic consistency, and structural fidelity. This fully validates the effectiveness of the designed framework in multimodal semantic modeling and structure-aware generation. By jointly modeling discriminative text-image relationships and guiding the generation process with fine-grained structural priors, the model is capable of precisely addressing both "what to generate" and "how to generate," showing strong potential for building high-fidelity and controllable image generation systems.

This paper also gives the impact of different text embedding dimensions on the image-text alignment performance, and the experimental results are shown in Figure 2.

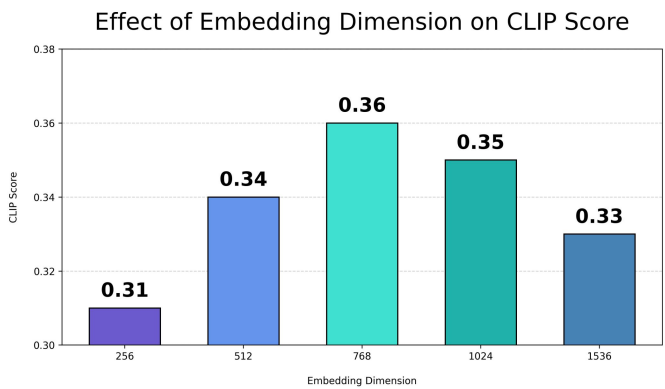

Figure 2. The impact of different text embedding dimensions on image-text alignment performance

Ablation results reveal a marked dependence of text-embedding dimensionality on cross-modal alignment: as the embedding size increases from 256 to 768, the CLIP Score rises monotonically, peaking at 0.36 at 768 dimensions, where fine-grained semantics are most faithfully captured and preserved during structure-guided synthesis; beyond this point (1024 and 1536 dimensions) the score declines slightly, indicating that excessive representational capacity introduces redundant cues that dilute structural guidance and impair text-image coherence. These findings demonstrate that judicious dimensionality selection is pivotal for balancing expressive power against overfitting, thereby reinforcing the synergy between the proposed contrastive mechanism and structural priors and providing a robust foundation for high-fidelity, semantically consistent image generation (see Figure 3 for full results).

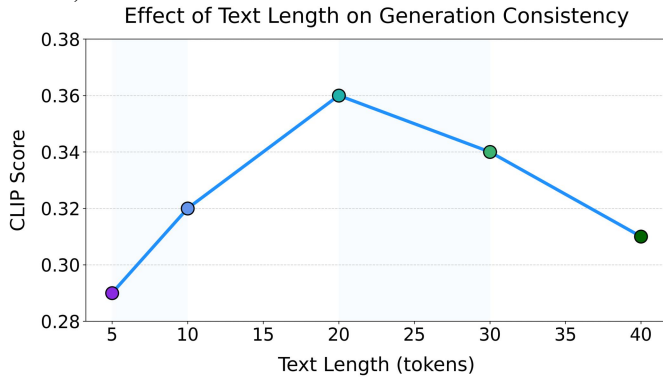

Figure 3. The impact of text description length variation on generation consistency

Ablation experiments demonstrate a clear, non-linear relationship between textual description length and cross-modal alignment quality. As the prompt expands from 5 to 20 words, the CLIP Score rises steadily—reaching its apex at 20 words—because a moderately detailed statement supplies sufficient structural and semantic cues for the contrastive module to learn discriminative features and for the structure-aware generator to synthesise images that faithfully reflect the intended content. In contrast, very short prompts ($\approx 5$ words) lack the semantic richness required for the model to disambiguate fine-grained meaning, resulting in incomplete reconstructions and diminished alignment. Conversely, extending the input to 30 – 40 words introduces semantic redundancy and peripheral information; this surplus diverts attention from core concepts, weakens the influence of structural guidance, and ultimately lowers the CLIP Score. These findings underscore that, within a framework combining structural priors and contrastive objectives, careful calibration of prompt length is essential: an intermediate span strikes the optimal balance between expressive capacity and cognitive load, thereby maximising both generation fidelity and text–image consistency.

## VI. CONCLUSION

This paper addresses the long-standing challenges of inaccurate semantic alignment and insufficient structural consistency in text-to-image generation. A high-fidelity generation method is proposed by integrating text-image contrastive constraints with structural guidance mechanisms. By introducing a discriminative contrastive learning framework, the model significantly improves the precision and stability of cross-modal semantic representations. It also overcomes the limitations of traditional models in handling ambiguous text descriptions. At the same time, the structural guidance module acts as a spatial prior during the generation process. It plays a key role in enhancing the layout rationality and detail fidelity of the generated images. This method enables joint modeling of semantic expression and structural awareness, offering a more interpretable and controllable solution for multimodal generation tasks.

In experimental evaluations, the proposed method demonstrates strong performance across multiple key metrics. It validates the effectiveness of the proposed mechanisms in maintaining text-image consistency, visual quality, and structural fidelity. These capabilities enhance the practical usability of the generated images. They also provide a solid foundation for semantic control and fine-grained modeling in complex scenes, highlighting the practical value of structural constraints in generation tasks. From an application perspective, the proposed method shows broad potential across several critical domains. In virtual reality and intelligent interaction, the model can quickly generate semantically rich and structurally coherent image responses based on user input. It can also support high-quality content generation in fields such as assisted design, advertisement creation, and medical image enhancement. Furthermore, due to its controllability and interpretability, the method is well-suited for professional domains that require strict semantic consistency and structural regularity. These include remote sensing analysis, product appearance design, and scientific visualization. Such applicability further expands the scope and depth of generative models in real-world scenarios.

## VII. FUTURE WORK

In future research, more fine-grained structural guidance mechanisms can be explored. Examples include semantic layouts, scene graphs, or graph-based priors, which can enhance spatial logic and semantic consistency in the generated content. With access to large-scale cross-domain multimodal corpora, dynamic text understanding and multi-stage generation strategies can be introduced. This would improve the model's generalization and reasoning capabilities for complex descriptions. In addition, integrating multi-task learning frameworks with tasks such as object detection, segmentation, and caption generation may lead to a unified generative system with cognitive ability and controllable expression. This could provide a more solid technical foundation for the next generation of intelligent multimodal generation systems.